\documentclass{llncs}

\usepackage{helvet} 
\usepackage{courier}  
\usepackage{url}
\usepackage{graphicx}  

\usepackage{grffile}

\usepackage{times}

\usepackage{wrapfig}

\usepackage[small]{caption}
\usepackage{subcaption}
\usepackage{amsmath,amssymb}

\usepackage{algorithm,algpseudocode}

\newcommand{\dmnd}[2]{\hfill\parbox[t]{\dimexpr #1\linewidth-\algorithmicindent}{$\triangleright$ #2\strut}}

\algnewcommand{\IfThen}[2]{
  \State \algorithmicif\ #1\ \algorithmicthen\ #2}

\algnewcommand{\IfThenElse}[3]{
  \State \algorithmicif\ #1\ \algorithmicthen\ #2\ \algorithmicelse\ #3}

\newcommand{\ffrac}[2]{\ensuremath{\frac{\displaystyle #1}{\displaystyle #2}}}

\newcommand{\diffsatshort}{$\partial$SAT/ASP}
  
\usepackage{listings}
\usepackage[usenames,dvipsnames]{xcolor}
\newcounter{definitionnumber}

\DeclareMathOperator*{\argmin}{argmin}

\algblock{ParFor}{EndParFor}

\algnewcommand\algorithmicparfor{\textbf{parfor}}
\algnewcommand\algorithmicpardo{\textbf{do}}
\algnewcommand\algorithmicendparfor{\textbf{end\ parfor}}
\algrenewtext{ParFor}[1]{\algorithmicparfor\ #1\ \algorithmicpardo}
\algrenewtext{EndParFor}{\algorithmicendparfor}



\makeatletter
\algrenewcommand\ALG@beginalgorithmic{\small}
\makeatother

\begin{document}
 	
\title{Differentiable Satisfiability and Differentiable Answer Set Programming for Sampling-Based Multi-Model Optimization\thanks{Extended and revised version of a paper \cite{DBLP:conf/ilp/Nickles18a} in Proceedings of the 5th International Workshop on Probabilistic Logic Programming (PLP'18)}}

\author{Matthias Nickles}

\authorrunning{M. Nickles}

\institute{National University of Ireland, Galway\\School of Engineering and Informatics\\
	\email{matthias.nickles@nuigalway.ie} }

\maketitle
\begin{abstract}
We propose Differentiable Satisfiability and Differentiable Answer Set Programming (Differentiable SAT/ASP) for multi-model optimization. Models (answer sets or satisfying truth assignments) are sampled using a novel SAT/ASP solving approach which uses a gradient descent-based branching mechanism. Sampling proceeds until the value of a user-defined multi-model cost function reaches a given threshold. As major use cases for our approach we propose distribution-aware model sampling and expressive yet scalable probabilistic logic programming. As our main algorithmic approach to Differentiable SAT/ASP, we introduce an enhancement of the state-of-the-art CDNL/CDCL algorithm for SAT/ASP solving. Additionally, we present alternative algorithms which use an unmodified ASP solver (Clingo/clasp) and map the optimization task to conventional answer set optimization or use so-called propagators. We also report on the open source software DelSAT, a recent prototype implementation of our main algorithm, and on initial experimental results which indicate that DelSAT's performance is, when applied to the use case of probabilistic logic inference, on par with Markov Logic Network (MLN) inference performance, despite having advantageous properties compared to MLNs, such as the ability to express inductive definitions and to work with probabilities as weights directly in all cases. Our experiments also indicate that our main algorithm is strongly superior in terms of performance compared to the presented alternative approaches which reduce a common instance of the general problem to regular SAT/ASP.

\end{abstract}

\section{Introduction}
\label{sec:intro}

Modern SAT and Answer Set solvers are, like their closely related cousins constraint processing and Satisfiability Modulo Theories (SMT), powerful and fast tools for logical reasoning. We present an approach which utilizes and enhances current SAT/ASP solving algorithms for sampling and multi-model optimization, with probabilistic inference and distribution-aware witness sampling as the focused - but not the only - use cases. We build on previous work \cite{Nickles2018b,Nickles2018a} and add new algorithms, in particular methods which require only an existing, unmodified ASP solver or map the task to a regular answer set optimization problem. Compared with \cite{DBLP:conf/ilp/Nickles18a}, in addition to several minor revisions we also report on a new open source implementation of our main algorithm (\textit{DelSAT}) and we have enhanced the experimental evaluation section.\\
With \textit{multi-model optimization} we denote the search for a multi-set of models (satisfying Boolean assignments respectively answer sets) which indirectly represents an (approximate) minimum of a user-provided cost (loss) function. We consider cost functions defined over certain statistical properties of the model multi-set, namely frequencies of atoms (propositional variables), with cost functions over fact, rule, clause or entire model weights as straightforward instances, and we incrementally sample models until a cost minimum is reached, with partial derivatives of the cost function guiding the assignment of decision literals in the SAT or ASP solving process. At this, the algorithms proposed in this work do not impose any restrictions (such as a probabilistic independence requirement) on the variables involved.\\
In the use case of probabilistic logic programming (a form of declarative probabilistic programming), the resulting multi-model optimum approximates a probability distribution over possible worlds (models) induced by probabilistic constraints (encoded as the cost function) and non-probabilistic rules and clauses (a regular ASP program or Boolean formula in CNF of which all sampled models are answer sets respectively satisfying assignments). Probabilistic rules can be handled using simple syntactic sugar. By sampling only as many models as required for cost minimization, we reduce the number of expensive conventional deductive inference steps and avoid the combinatorial explosion of materialized possible worlds with increasing number of nondeterministic atoms which typically precludes the use of straightforward optimization techniques (such as linear programming) in probabilistic logic programming.\\ 
In principle, arbitrary differentiable cost function can be used (although obviously not all cost functions lead to convergence of the optimization process) and there are no restrictions on rules or clauses, or the random variables (such as independence assumptions), except consistency. The expressiveness of the framework is thus quite high, and, despite the use of sampling, computation times remain a challenge (in particular in comparison with approaches which deliberately put restrictions into place, such as frameworks based on the Distribution Semantics). We tackle this concern with our first concrete algorithm (a variant of the algorithm presented in \cite{Nickles2018b}) by integrating the cost minimization steps directly into a state-of-the-art ASP/SAT solving approach \cite{gekanesc07a}, and we compare its performance for the major use case of probabilistic logic programming experimentally with that of several alternative methods introduced in this paper as well as with a scalable software for Markov Logic Networks.\\
The remainder of this paper is organized as follows: The following section introduces basic concepts and notation. Sect. \ref{diffsat} presents our general approach and proposes several concrete computation methods. Sect. \ref{sec:experiments} presents results from preliminary experiments, and Sect. \ref{sec:relatedWork} discusses related approaches. Sect. \ref{sec:conclusion} concludes.

\section{Preliminaries}
\label{notation}

We consider ground normal logic programs under stable model semantics and SAT problems in the form of propositional formulas in Conjunctive Normal Form (CNF). Recall that a normal logic program is a finite set of rules of the form\\
$h\ {:}{-}\ b_1, ..., b_m,\ not\ b_{m+1}, ..., not\  b_n$ (with $0 \le m \le n$). $h$ and the $b_i$ are atoms without variables. $not$ represents default negation. Rules without body literals are called facts. Most other syntactic constructs supported by contemporary ASP solvers (like integrity constraints, choice rules or classical negation) can be translated into (sets of) normal rules. 
We consider only ground programs in this work and all atoms and literals are assumed to be ground.
The \textit{answer sets} (\textit{stable models}) of a normal logic program are as defined in \cite{stable-model-semantics}. Throughout the paper, we use the term ``answer set program'' to mean a ground normal logic program and ``model'' in the sense of answer set or, in the SAT case, a complete truth assignment such that the formula evaluates to true, however, to use the same model notation with both ASP and SAT, we generally do not show false atoms in models, i.e., a model is represented as a set of atoms which hold in the model. As common in probabilistic ASP, we identify\textit{ possible worlds} with models. $\Psi_\mathit{\Upsilon}$ denotes the set of all answer sets or satisfying assignments of answer set program or propositional formula $\Upsilon$. Sometimes we use only logic programming terminology where the translation to SAT terminology is obvious (e.g., ``program'' instead of set of clauses). The set of all atoms respectively propositional variables in a program or formula $\Upsilon$ is denoted as $\mathit{atoms}(\Upsilon)$. We write $\overline{S}$ to denote a set of negative literals $\{ \overline{s_i}: s_i \in S\}$.\\
A \textit{partial assignment} denotes an incomplete ``model under construction'': a sequence of literals which have been iteratively added (assigned) to the assignment (and sometimes retracted from the assignment in backtracking steps) by the SAT or ASP solver until the assignment is complete or the procedure aborts in case of unsatisfiability. \\
In our main algorithm (Sect. \ref{Diff-CDNL-SAT/ASP}) which is based on CDNL (Conflict-Driven Nogood Learning) \cite{gekanesc07a}, we use a unified approach to SAT and ASP solving and \textit{nogoods} which correspond to clauses in CNF but with all literals negated; a concept originally introduced for constraint solving. Clauses and rules are translated into nogoods in a preprocessing step (covering Clark's completion in the ASP case). Additional nogoods are learned from conflicts and loops (in non-tight ASP programs). The use of nogoods instead of clauses is not essential for our approach, creating a variant based on the similar but older CDCL (Conflict-Driven Clause Learning) approach instead of CDNL would be trivial.

\subsection{Cost Functions and Parameter Atoms}

The cost functions considered in this work are user-provided functions of several variables. Each variable corresponds to a so-called \textit{parameter atom}. When evaluating the cost function, we instantiate each variable with the normalized count (\textit{frequency}) of the respective parameter atom in a possibly incomplete \textit{sample}. The \textit{sample} is a multi-set of models sampled with replacement from the complete set $\Psi_\mathit{\Upsilon}$ of answers sets of the given program, respectively the set of all satisfying assignments (SAT case). Where a parameter atom or its negation can occur we speak of \textit{parameter literals}. Parameter atoms can occur in rules and clauses of the given answer set program or Boolean formula without limitations, and may even be conditioned on the truth values of other atoms, including other parameter atoms. The set of parameter atoms is denoted as $\theta$ or $\{\theta_i\}$, and $\theta_i$ is the i-th parameter atom under some arbitrary but fixed order. We denote the individual cost function variables corresponding to the parameter atoms as $\theta_i^v$ or $a^v$ (where $a$ is a parameter atom). The parameter atoms should be chosen by the user from the overall set of atoms in the program or set of clauses (theoretically, simply \textit{all} atoms (variables) could be declared parameter atoms, with ``normal'' (non-probabilistic) atoms just not appearing in cost functions - the algorithms would still work then, but this might be very inefficient.) \\
With each newly sampled model, the frequencies of the parameter atoms are updated as follows.  $\beta(\mathit{sample}) = \langle\beta^\mathit{sample}(\theta_1), \beta^\mathit{sample}(\theta_2), ...\rangle$ is defined as the (parameter) frequencies vector {\small
$  \langle \ffrac{|[m_j:\ m_j \in \mathit{sample}, \theta_1 \in m_j]|}{|\mathit{sample}|}, ...,$
$\ffrac{|[m_j:\ m_j \in \mathit{sample}, \theta_n \in m_j]|}{|\mathit{sample}|} \rangle$ 
} of parameter atom frequencies in the model multi-set $\mathit{sample}$. $\beta^\mathit{sample}(\theta_i)$ denotes the frequency of parameter atom $\theta_i$ in $\mathit{sample}$. We omit $\mathit{sample}$ and simply write $\beta(\theta_i)$ where it is clear from the context what the sample is. $\mathit{cost}(\beta(\theta_1), \beta(\theta_2), ...)$ evaluates the cost function $cost$ over parameter frequencies vector $\beta(\mathit{sample})$. We sometimes write $\mathit{cost}(\mathit{sample})$ in place of $\mathit{cost}(\beta(\theta_1), \beta(\theta_2), ...)$.\\
It makes sense to allow only parameter atoms whose truth values are not fully fixed by the input program or formula. To ensure this in the ASP case, we can give the logic program the shape of a so-called \textit{spanning program} \cite{NicMil15,ca7613df8f3c4bc591342bc8391977ae} where uncertain atoms $a$ are defined by \textit{spanning formulas}: choice rules or analogous constructs amounting to nondeterministic facts $0\{a\}1$ or (informally) $a \vee not\ a$. 
However, our framework does not require any particular form of the input program or formula.\\
Informal examples for cost functions are ``In 30\% of all models, atom $a$ should hold and in 40\% of all models, atom $b$ should hold'' or ``Atom $a$ should hold more often than the square root of the frequency of atom $b$ minus 10\%''. In principle, other types of cost functions which refer to other properties of the current sample multi-set are conceivable too, provided the model sampler is able to minimize such a cost. 

\subsection{Cost Functions and Parameter Atoms for Distribution-Aware Model Sampling and Probabilistic Inference }
\label{sec:linear}

For the application cases of model sampling and deductive probabilistic inference, cost functions specify probabilistic constraints, whereas the plain ASP rules or SAT clauses serve as hard logical constraints. The sampling process then generates models until the overall sample (i.e., the multi-set of models) satisfies all these constraints, up to the user-specified cost function threshold.  Optionally, sampling can then continue until some requested number of models have been sampled while the cost is equal or below the threshold (this can be useful, e.g., to increase the entropy of the sample). \\

To perform marginal inference, a subsequent step calculates the probabilities of query atoms in the usual way as the frequencies of these atoms in the sample, identifying the sampled models with possible worlds. For MAP inference, select the most frequent model in the sample and project it to the query.\\

 More concretely, we consider a setting where the probabilistic constraints are provided by the user in form of \textit{weights} associated with individual parameter atoms $\theta_i$. Weights directly represent probabilities. Weights can also be attached to entire models (in the form of conjunctions, which is straightforward) or to arbitrary rules, by introducing weighted fresh auxiliary atoms as ``shortcuts'' for these rules, using the following scheme:  $h\ {:}{-}\ b_1, ..., b_n,\ not\ \mathit{aux}$ and $\mathit{aux}\ {:}{-}\ b_1, ..., b_n,\ not\ h$ \cite{ca7613df8f3c4bc591342bc8391977ae}.\\
As one (but not the only) suitable cost function which can be derived directly from a set of weighted parameter atoms, we propose the use of the \textit{Mean Squared Error} (MSE) $cost(\theta_1^v, \theta_2^v, ...) := {\large \frac{1}{n}}\sum_{i=1}^n(\beta(\theta_i) - \phi_i)^2$ (squared Euclidean distance normalized with the number $n$ of parameter atoms). The $\phi_i$ are the user-defined weights of the parameter atoms $\theta_i$. \\
With this cost function, appending models to the sample until the cost reaches zero (i.e., maximum accuracy) corresponds to finding a solution of a linear equation system with additional conditions to ensure that the solutions form a probability distribution, with the probabilities of \textit{all} possible worlds $\Psi_\mathit{\Upsilon}$ as the unknowns and the actual model frequencies in $\mathit{sample}$ as approximate solution vector (details are provided in \cite{Nickles2018b}). Queries can then be performed as usual, by adding up the approximated probabilities of those possible worlds where the query holds (that is, the frequencies of those models in $\mathit{sample}$ which positively contain the query atoms). \\ 
As an example for how to assign probabilities to atoms using cost functions, consider function $\mathit{cost}(a^v, b^v) = ((0.2 - a^v)^2 + (0.6 - b^v)^2)/2$) which specifies that the weight of parameter atom $a$ should be 0.2 and the weight of parameter atoms $b$ should be 0.6. \\ 
As another example, this time for a cost function different from MSE (but also solvable using Differentiable ASP), consider $\mathit{cost}(\mathit{aux}^v, q^v) = (0.4 - \mathit{aux}^v / q^v)^2$ which specifies that the conditional probability $Pr(p|q)$ is 0.4. The accompanying answer set program needs to include rules $\mathit{aux}\ {:}{-}\ p, q$ and $p\ {:}{-}\ \mathit{aux}$ and $q\ {:}{-}\ \mathit{aux}$. 

\section{Differentiable SAT/ASP}
\label{diffsat}

To find a sample of models which minimizes the cost function, our approach iteratively appends models to multi-set $\mathit{sample}$ until the cost falls below a specified threshold (allowing the user to trade speed against accuracy). All models are answer sets or satisfying truth assignment of the given answer set program or Boolean formula, ensuring that they adhere to the given ``hard'' logical constraints. Partial cost function derivatives guide, during the generation of each individual model, the SAT/ASP solving process on the level of branching decisions, namely truth value assignments to parameter atoms. 
\begin{figure}[h]
\vspace*{-0.8cm}
\centering
\includegraphics[width=1.05\linewidth,height=6.9cm]{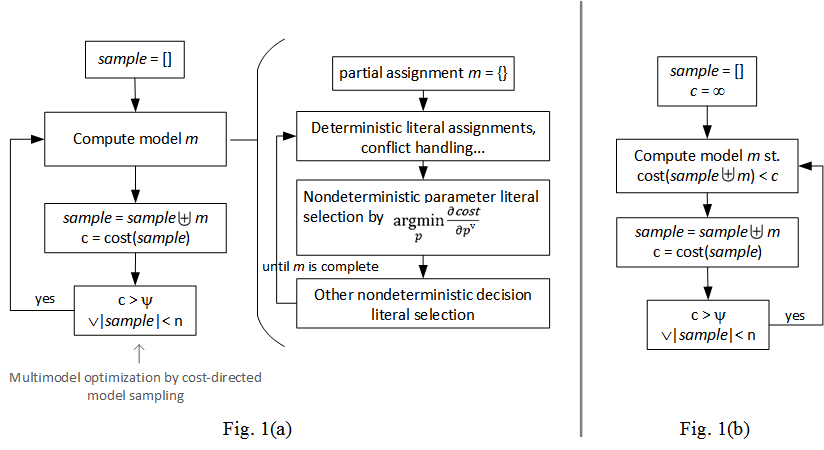}
\caption{Differentiable SAT/ASP (outline)}
\label{fig:overalloverall}
\vspace*{-0.8cm}
\end{figure}

Fig. \ref{fig:overalloverall}(a) shows a high level view of this approach, named \textit{Differentiable SAT/ASP} or \diffsatshort\ for short. An outer loop (left side of Fig. \ref{fig:overalloverall}(a)) samples models and adds them to an initially empty multi-set $\mathit{sample}$ until the termination criterion is reached (there are several specific possibility for checking termination, depending on the nature of the cost function and the use case: if the cost expression is not too large, we can check if it is equal or below a given threshold $\psi$ (accuracy), and/or we could perform a stagnation check on the cost or the parameter atom frequencies. In some applications it might also be sensible to demand a minimum sample size $n$ in addition to reaching threshold $\psi$, e.g., to increase the sample entropy). In our experiments, we simply stopped sampling when the cost reached or fell below $\psi$. \\
The models are sampled with the aim of reducing the multi-model cost, using a form of discretized gradient descent. We approach this with a special branching rule for selecting decision literals (literals not enforced by propagation) for inclusion in the partial assignment: Each time the solver nondeterministically (i.e., not forced by rules or clauses) extends the partial assignment with a not yet assigned literal, the literal is selected from all unassigned parameter literals (if any) if the value (or its negation in case of negative literals) of the cost function's partial derivative with respect to this literal is minimal (compared with the values obtained for the other unassigned parameter literals). Since the parameter search space is discrete, we could theoretically measure the cost impacts of hypothetically assignments of candidate literals directly. But taking the partial derivatives with respect to the parameter atoms splits the overall cost calculation (which might be complex) into typically simpler calculations whose results can even be pre-computed and ranked after each new model according to their current values, after updating the frequencies vector $\beta$ with the new model. Finally, for branching decisions on non-parameter decision literals, some conventional branching heuristics (e.g., BerkMin) can be used. \\
Fig. \ref{fig:overalloverall}(b) outlines a variant where each new model is explicitly required to lower the cost, without specifying how to achieve this. A specific approach to this variant, proposed in \cite{Nickles2018b}, uses a so-called \textit{cost backtracking} mechanism which in case a model candidate fails to improve the cost, jumps back to the most recent literal decision point with untried parameter literals and tries a new parameter literal. 
\cite{Nickles2018b} also indicates how cost backtracking and a split of the set of parameter atoms into \textit{measured atoms} and actual parameter atoms can be utilized for inductive weight learning and abductive reasoning (which are not possible using the approach in Fig. \ref{fig:overalloverall}(a)). \\
In the following subsections, we propose concrete approaches to put the general approach in Fig. \ref{fig:overalloverall}(a) into practice. 

\subsection{Implementing Differentiable SAT/ASP based on CDNL }
\label{Diff-CDNL-SAT/ASP}

A concrete approach to the rather general scheme described above is to enhance the current state-of-the-art approach to Answer Set Programming (Conflict-Driven Nogood Learning (CDNL) \cite{gekanesc07a}) or a similar approach (CDCL (Conflict-Driven Clause Learning)- or DPLL-style solving) with a new branching rule which selects free parameter literals for inclusion into the current partial assignment according to their negative ``impact'' on the cost, determined using partial derivatives wrt. parameter atoms. This approach, which we call \textit{Diff-CDNL-ASP/SAT} (as it covers both SAT and Answer Set solving), is shown as Algo. \ref{algo:OptiSAT-CDNL-ASP-short}. The SAT solving path through the algorithm, enabled by parameter $\mathit{SATmode}$, is largely identical to the more complex ASP path, with stable model checks and loop handling omitted. The algorithm is a variant of the approach presented in \cite{Nickles2018b}, with a somewhat more general branching approach using partial derivatives but omitting the optional cost backtracking.\\
Algo. \ref{algo:OptiSAT-CDNL-ASP-short} iteratively adds literals to a partial assignment until all literals are covered and no \textit{nogood} is violated. Important steps are unit propagation, i.e., the deterministic expansion of the partial assignment using Boolean constraint propagation (procedure \textsc{BCP}) with literals ``fired'' by nogoods, and conflict analysis (conflict means at least one nogood is subset of the partial assignment) and handling, including the deriving of further nogoods from the analysis of conflicts and backjumping (undoing recent literal assignments) in line 21 (details follow \cite{gekanesc07a} and are omitted here for lack of space).\\
The procedure for generating a single model (the while-loop from line 6) is thus guided by the following factors: 1) the initial set of nogoods obtained from the CNF clauses or Clark's completion of the answer set program, 2) further nogoods added to this set by conflict handling or due to the presence of loops (ASP mode), 3) the given cost function and set of parameter atoms, and 4) our new branching approach for assigning parameter literals (lines 9 to 13). The inner while loop ends once all atoms are covered as positive or negative literals (or UNSAT). Afterwards (line 25), the stable model check takes place (unless in SAT mode), and the new model is appended to the multi-set $\mathit{sample}$. The outer loop (from line 3) ends when a convergence criterion is met (e.g., when the cost falls below the given accuracy threshold $\psi$ (line 32) and we have obtained the requested number of models, or if there is no more progress).\\
The decision branching rule is the main different to regular CDNL: In lines 11ff., we select the next parameter literal according to the previously described approach using partial derivatives. At this, it was in all our initial tests sufficient for reaching convergence to fix a new ranking of parameter literals by the values of the respective partial derivatives only after a new model was generated (ignoring the current partial assignment in the computation of the parameter atom frequencies) and to use this ranking to determine the next decision literal in line 11.\\
For further details on the non-probabilistic aspects of the algorithm, we need to refer to \cite{gekanesc07a} for lack of space. Note that for loop handling, Algo. \ref{algo:OptiSAT-CDNL-ASP-short} uses the older and simpler ASSAT approach \cite{LIN2004115}, just to simplify our initial implementation.

\begin{algorithm}[ht!]
	\caption{Diff-CDNL-ASP/SAT}
	\label{algo:OptiSAT-CDNL-ASP-short}
	\begin{algorithmic}[1]  \raggedright
		\State Arguments: $\Upsilon$ (program or formula), $nogoods(\Upsilon)$ (initial set of nogoods, derived from given clauses or answer set program), $\psi$ (accuracy), $\theta$, $\mathit{cost}$, $\mathit{SATmode}$, $n$ (minimum number of models sampled with specified accuracy, 0 in our experiments)
		
		\State $\mathit{sample} \gets [\ ]$
		
		\Repeat  \Comment{Outer loop (enhances multi-set $sample$)}
		
		\State $as \gets [\ ]$  \Comment{Assignment over literals}
		
		\State $dl \gets 0$  \dmnd{0.5}{Decision level $dl$ initially 0 (no nondeterministic decisions made yet)}
		
		\While{$\mathit{incomplete}(as) \vee \mathit{conflicting}(as)$}  \Comment{Inner loop (computes a single model)}
		
		\State $as \gets \textsc{BCP}(\{\}, \mathit{nogoods(\Upsilon)})$ \dmnd{0.5} { Iterated unit propagations (enhancing  $as$ with unit-resulting literals until fixpoint). }
		
		\If{$\neg \mathit{conflicting}(as) \wedge \mathit{incomplete}(as)$}		\Comment {Branching...}
		
		\State $dl \gets dl + 1$	
		
		\State $ua\theta \gets (\theta \cup \overline{\theta}) \backslash as$ \Comment unassigned parameter literals (decision literal candidates)
		
		\If{$ua\theta \neq \{\}$}
		
		\State	$decLit \gets \argmin_{p \in ua\theta}  \mathit{sg}\ \cdot  {\frac{\partial  \mathit{cost}}{\partial pa^v}}(\beta^\mathit{sample}(\theta_1), \beta^\mathit{sample}(\theta_2), ...)$, 
		\State	\hspace{1.4cm} with $(pa, \mathit{sg}) = \begin{cases} 
		
		(p, 1)\  \mathit{ if }\  p \in \mathit{atoms(\Upsilon)}\\
		(\overline{p}, -1)\  \mathit{ otherwise }
		
	\end{cases}
	$
	
	\State $decLitPr \gets 1 - \mathit{noise}$   \Comment $\mathit{noise} \ll 1$
	\Else		
	\State $(decLit, decLitPr) \gets ...$ (using some conventional branching heuristics)  
	\EndIf
	
	\State \textbf{if} $rand_0^1 < decLitPr$ \textbf{then} $as \gets as \cup decLit$ \textbf{else} $as \gets as \cup \overline{decLit}$ 
	
	
	\Else \If{$\mathit{conflicting}(as)$}
	
	\State ConflictHandling with nogood learning, back jumping (or stop with UNSAT  
	\State if $dl = 0$), adding negated UIP to assignment (as in plain CDNL \cite{gekanesc07a})				
	
	\EndIf
	
	\EndIf
	
	\EndWhile  
	
	\State $\mathit{modelCand} \gets $ positive literals in $as$ (except body literals)  \Comment (see Sect. \ref{notation})
	
	\If{$\mathit{SATmode}$ $\vee\ \mathit{tight}(\Upsilon)\ \vee\ stable(\mathit{modelCand})$}
	
	\State $\mathit{sample} \gets \mathit{sample} \uplus \{ \mathit{modelCand} \}$
	
	
	\Else \If{$\neg \mathit{SATmode}$}
	\State add proper subset of loop nogoods	 \Comment{(or some other approach to non-tight programs)}
	\EndIf
	\EndIf
	
	\Until {$cost(\beta^\mathit{sample}(\theta_1), \beta^\mathit{sample}(\theta_2), ...) \le \psi\ \wedge$ $|\mathit{sample}| \ge n$ }  \hspace{0.4cm} (or until cost stagnation or maximum number of trials exceeded)
\end{algorithmic}
\end{algorithm}	
	
\subsection{DelSAT}
\label{DelSAT}
	
We have implemented an optimized, enhanced variant of Algorithm \ref{algo:OptiSAT-CDNL-ASP-short} as the open source SAT/ASP solver \textit{DelSAT} \footnote{\texttt{https://github.com/MatthiasNickles/DelSAT}}. As for its regular SAT and ASP solving capability, DelSAT is a parallel CDNL-style portfolio solver which allows to let various parameterizations of the solver compete against each other in multiple parallel solver threads. Each time the solver generates a single model, a user-defined number of solver threads are started, each with a different solver configuration and/or different randomizations of certain data structures. The fastest configuration' (i.e., where the respective solver thread completes first) is then used for all subsequent model generations in the overall sampling process.\\

DelSAT is programmed largely in Scala (and a small amount of Java) and runs on the Java Virtual Machine (Java 8 or higher), making it usable in a large number of computing environments. It accepts input in the form of DIMACS-CNF, DIMACS-CNF with cost functions, aspif (ASP intermediate language) or aspif with cost functions, and can thus be used as a plain SAT or Answer Set solver as well as a sampling and optimization tool. Used as an Answer Set solver, DelSAT supports ground normal as well as disjunctive programs. Non-ground or AnsProlog input needs to be preprocessed. Further details can be found on the GitHub page of DelSAT. 
	
\subsection{Differentiable ASP using propagators (\textit{Diff-ASP-ThProp})}
\label{Diff-ASP-ThProp}

After having presented our main contribution in the two previous sections, we also report alternative approaches which, while not being particularly fast, can be realized using an unmodified existing ASP or SAT solver. As the first of these alternative approaches, we make use of Clingo's\footnote{\texttt{https://github.com/potassco/clingo}} \textit{propagators}. We show how to do this using preliminary code\footnote{{\footnotesize \url{https://github.com/MatthiasNickles/Diff-ASP-Propagators}} }
 (file {\footnotesize \verb#propdiff_1.py.lp#} under {\footnotesize \url{https://github.com/MatthiasNickles/Diff-ASP-Propagators}}), instantiated with an MSE-shaped example cost function and two parameter atoms $a$ (with given probability 0.6) and $b$ (probability 0.2). It requires only Clingo (tested with version 5.2) with Clingo's Python scripting interface and Python 2.7.\\
While custom propagators cannot directly implement a branching heuristics, they can be used to add (parameter) literals in form of singleton clauses to the ongoing solving process, intercepting propagation. We compute the parameter literal we would like to assign next (again using the approach in Sect. \ref{diffsat}) and then pass this literal\\
 {\footnotesize (\verb#branch_param_lit#}) on to the propagator (lines 56ff. in the code) to add it to the  partial assignment.\\
The rest of the Python code is straightforward: The loop from line 155 corresponds to the outer loop in Algo. \ref{algo:OptiSAT-CDNL-ASP-short}: it iteratively calls the solver to compute new models, adds each newly sampled model to the model list {\footnotesize \verb#sample#} (in callback {\footnotesize \verb#on_model#}), updates frequencies and evaluates the cost (method {\footnotesize \verb#update_cost#}), and checks for convergence against threshold {\footnotesize \verb#psi#}.\\
The code supports both numerical and automatic differentiation (the latter using the \textit{ad} package\footnote{\url{https://pypi.org/project/ad/}}). In both cases,
the search for the parameter literal which gives the minimum partial derivative (i.e., steepest descent) is performed in lines 124ff.
Automatic differentiation wrt. parameter atoms takes place in method {\footnotesize \verb#__cost_ad#}. For numerical differentiation, we use a simple approximation which just adds a small value {\footnotesize \verb#h#} to the frequency of each respective parameter atom to estimate the
slope (method{\footnotesize  \verb#__cost_upd#}).\\
The actual cost function (including the given weights of the two parameter atoms) is in line 84 (we use the expression format of the ad-package for Python to represent the cost expression, also with numerical differentiation). \\ 
(Remark: We have also experimented with domain heuristics using Clingo's designated predicate {\footnotesize \verb#_heuristic/3#}, but found no way yet to make this reliably working for our use case. Also, this attempt appears to be much slower than all other approaches presented in this paper.)\\
\noindent An example for a simple associated background theory (file {\footnotesize  \verb#propdiff_bgk_1.lp#}) is 
{\vspace{-0.2cm} \scriptsize \begin{verbatim}
0{a}1. % spanning rule for parameter atom a 
0{b}1.
:- a, b. % an example for a hard rule
\end{verbatim} \vspace{-0.3cm}}

\noindent  The overall program is called with {\scriptsize \verb#clingo-python propdiff_1.py.lp propdiff_bgk_1.lp#}

\subsection{Direct cost minimization using model reification (\textit{Diff-ASP-Reification})}
\label{Diff-ASP-Reification}

As the final approach proposed in this paper, we use Clingo with reified predicates and models to solve the cost function directly (without derivatives involved), alternatively by mapping it to a conventional single answer set optimization task or by mapping the problem to ASP-encoded equation solving. \\
Here, each predicate in the original answer set program (not only the parameter atoms) whose truth value is not fully fixed is enhanced with a model number as extra argument. This way, we can let a single actual model (returned by Clingo) represent multiple reified models where each reified model consists exactly of those atoms in the actual model which share the same model number as their extra argument.\\
We distinguish two flavors of this approach: 1) Computation of one or more individually optimal answer set(s) using optimization statements or weak constraints (...:$\sim$...), as supported by several ASP solvers, including smodels, DLV and Clingo/clasp, and 2) using ASP directly for constrained linear equations solving. \\
We consider variant 1) first, shown in the code below for the same example as before (two uncertain atoms $a$ and $b$ with weights 0.2 and 0.6), and MSE as cost function format (adaptations to some other types of cost function should be straightforward). The number of reified models \verb|nmodels| does not need to be known precisely but should be at least $10^n$ where $n$ is the number of decimal places which should be accurate when using the overall result to query $\mathit{Pr}(a)$ and $\mathit{Pr}(b)$. We found this approach very slow in our preliminary experiments with default settings (more efficient encodings or optimization strategies might exist), but in any case it is useful to exemplify how our multi-model optimization task can be mapped to conventional answer set optimization using reification. 

{\scriptsize \vspace{-0.2cm} \begin{verbatim}
#const nmodels = 10.
model(1..nmodels).
mcount(0..nmodels).
{a(M)} :- model(M).  % spanning formulas
{b(M)} :- model(M).
:- a(M), b(M), model(M). % an example for a background knowledge rule (hard constraint)

wa(nmodels * 2 / 10).   % weight a = 0.2
wb(nmodels * 6 / 10). 	% weight b = 0.6
fa(F) :- F { a(M): model(M) } F, mcount(F).
fb(F) :- F { b(M): model(M) } F, mcount(F).

diffa(D) :- D = (W - F)**2, wa(W), fa(F). % alternatively: D = |F - W|
diffb(D) :- D = (W - F)**2, wb(W), fb(F).
#minimize { DA : diffa(DA) }.  %  minimize the distances betw. weights and frequencies
#minimize { DB : diffb(DB) }.

#show a/1.
#show b/1.
\end{verbatim} \vspace{-0.3cm}}

In variant 2) of our reification-based approach, we map the problem to a set of linear equations (or inequalities, if error tolerance bounds are considered) as outlined in Sect. \ref{sec:linear}, and encode it as a plain answer set program.  It is immediately clear that here the number of reified models introduces a bottleneck: Every predicate whose truth value is not fully fixed across all reified models needs to be reified, which multiplies the number of its instances with the overall number of reified models (\verb#nmodels# in the code below).  Nevertheless, as detailed in the next section, this simple approach fares surprisingly well for relatively small problem sizes, and significantly better than the approach using propagation (Sect. \ref{Diff-ASP-ThProp}). The following plain ASP program shows how to implement this for the example problem above.  

{\vspace{-0.2cm} \scriptsize \begin{verbatim}
#const tol = 3. % NB: tol has a different semantics than \psi
#const multiplier = 100.  % to map float numbers to integers; limits  precision
#const nmodels = 400.
model(1..nmodels).

wa(nmodels * 2 * multiplier / (10 * multiplier)).  % 2 represents given weight 0.2
W-tol < { a(M): model(M) } < W+tol :- wa(W).  
wb(nmodels * 6 * multiplier / (10 * multiplier)).  % 6 represents given weight 0.6
W-tol < { b(M): model(M) } < W+tol :- wb(W).

1{__aux_1(M);a(M)}1 :- model(M).  % spanning formulas
1{__aux_2(M);b(M)}1 :- model(M).

:- a(M), b(M).  % example for a hard background knowledge rule 

#show a/1.
#show b/1.
\end{verbatim} \vspace{-0.4cm}}

\section{ Experiments}
\label{sec:experiments}

To provide initial insight into the performance characteristics of the presented approaches in the domain of probabilistic inference, we have performed a number of experiments. Approaches considered were Tuffy 0.4 \cite{} (an inference and weight learning tool for Markov Logic Networks), DelSAT 0.2.1 (based on Diff-CDNL-ASP/SAT (Sect. \ref{Diff-CDNL-SAT/ASP})), Diff-ASP-ThProp (Sect. \ref{Diff-ASP-ThProp}) and Diff-ASP-Reification (Sect. \ref{Diff-ASP-Reification}), the latter in the equation solving variant (the version using \verb|#minimize| proved too slow with these experiments to be considered). Tuffy uses the MC-SAT algorithm for marginal inference. Tuffy has been selected as a scalable approach to MLN which is implemented in Java (and thus can use the same runtime environment as DelSAT). Markov Logic Networks and our approach have different semantics and input syntax, so comparative results should be considered with some caution. However, with DelSAT applied to inference in probabilistic logic, they share a similar use case, making the presented results useful to indicate the expected relative performance in this application domain from a potential users' point of view. Experiments differ from earlier works \cite{DBLP:conf/ilp/Nickles18a,Nickles2018b,Nickles2018a} mainly in the use of DelSAT instead of earlier significantly slower prototypical implementations.\\
 
\begin{figure}[h!]
	\centering
	\includegraphics[width=0.8\linewidth, keepaspectratio]{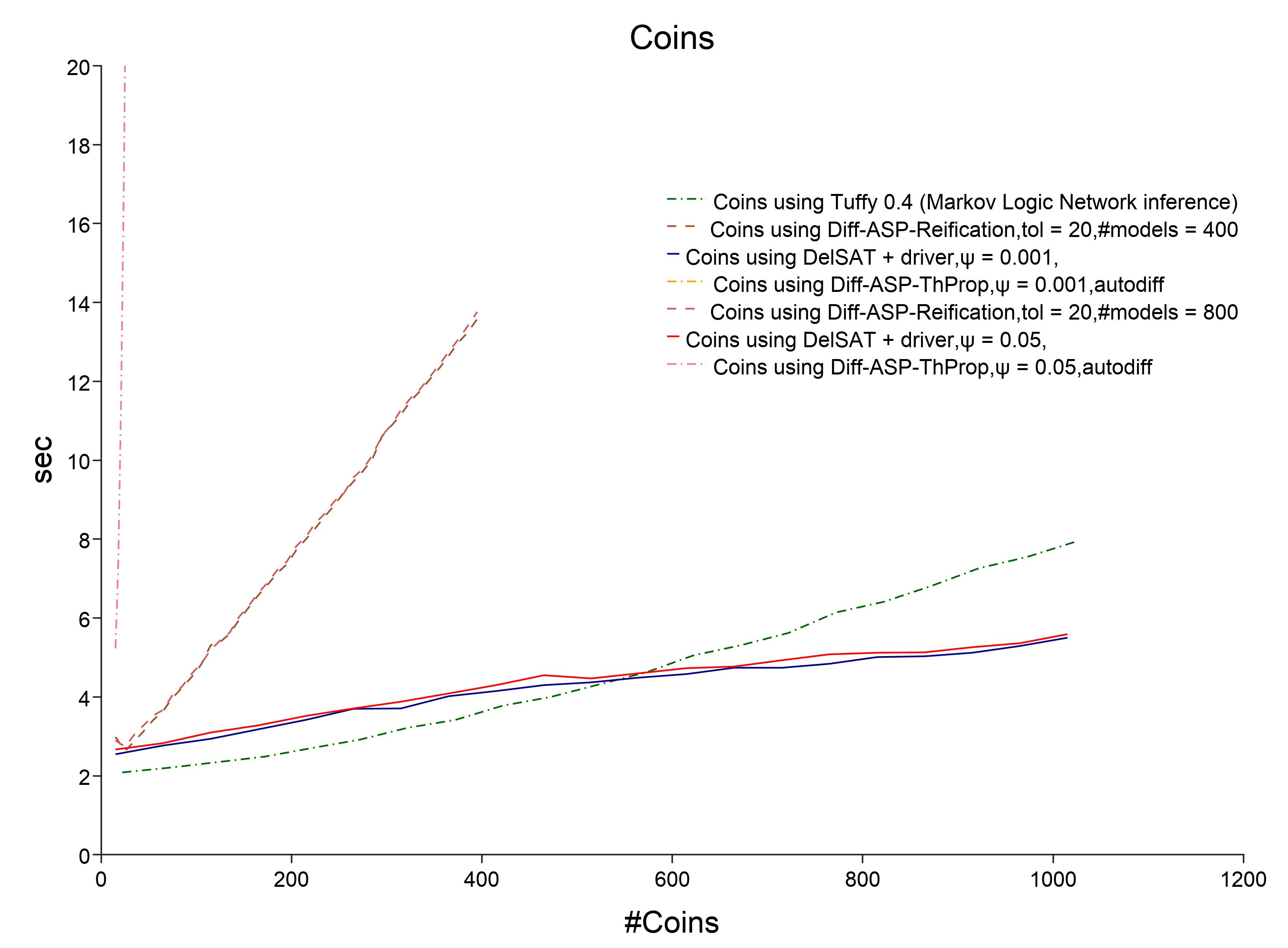}
	\caption{Coins Game}
	\label{fig:Coins}
\end{figure}

\begin{figure}[h!]
	\centering
	\includegraphics[width=0.8\linewidth, keepaspectratio]{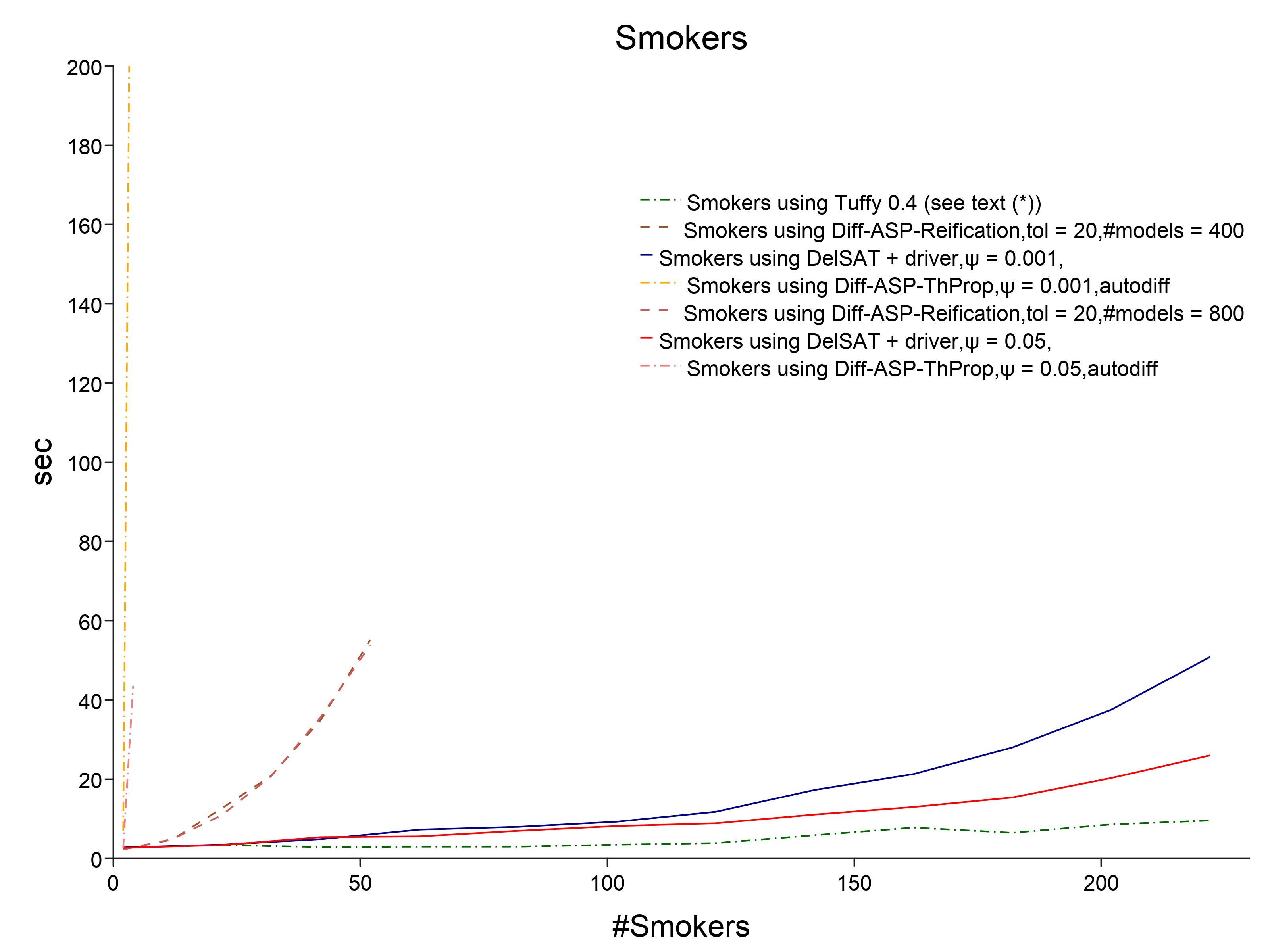}
	\caption{Friends \& Smokers}
	\label{fig:Smokers}
\end{figure}

\begin{figure}[h!]
	\centering
	\includegraphics[width=0.8\linewidth, keepaspectratio]{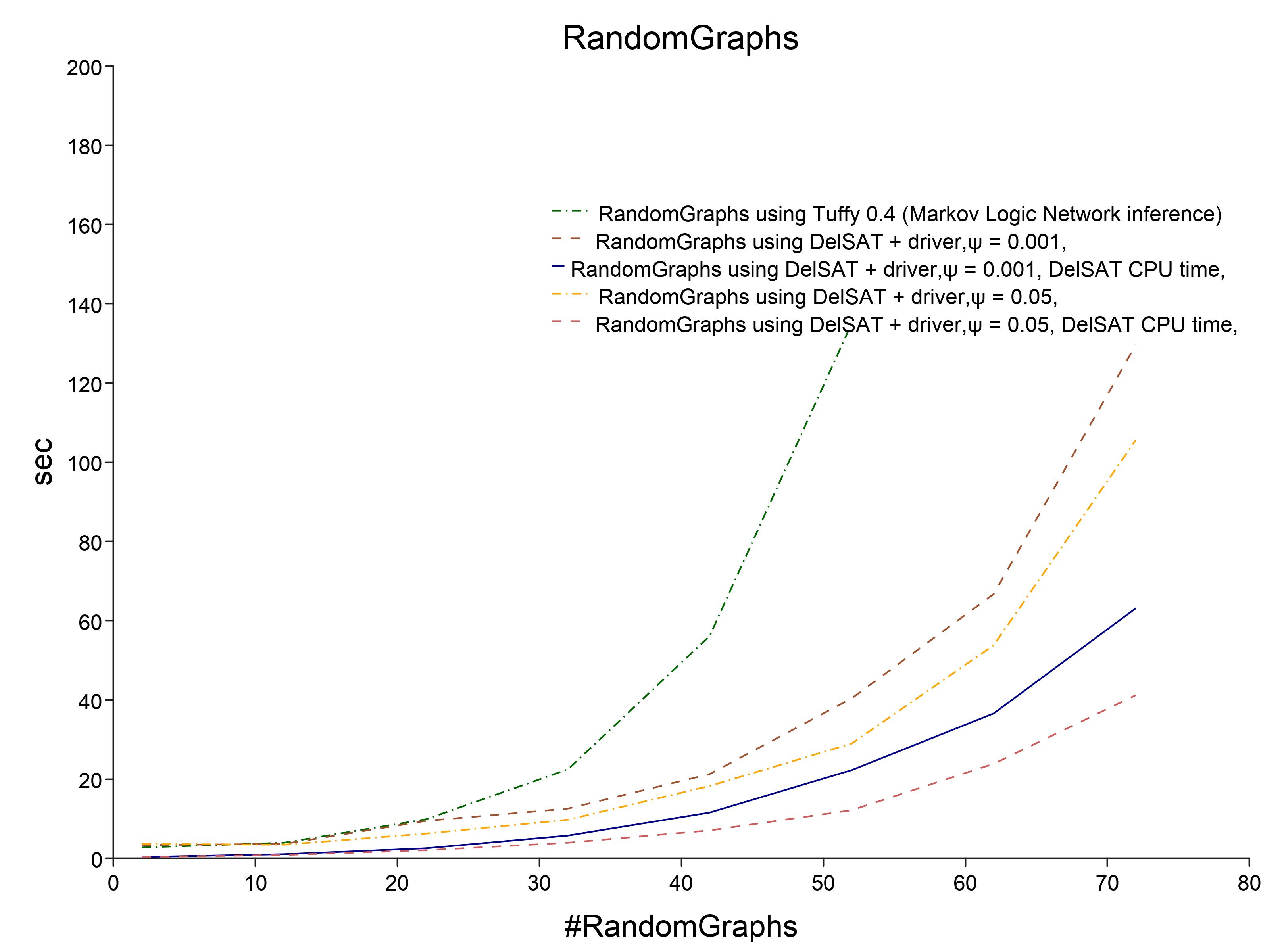}
	\caption{Random Graphs}
	\label{fig:RandomGraphs}
\end{figure}

We performed three synthetic experiments (Figs. \ref{fig:Coins}, \ref{fig:Smokers}, \ref{fig:RandomGraphs}): inference over a coin tossing game with background rules, inference over a version of the well-known ``Friends \& Smokers'' scenario (which exists in several variants in the literature) and inference over random graphs. \\

All times - unless where noted in the graph (``CPU time'') - are end-to-end (wall clock) times for invoking the respective tools as external programs and include some overhead for I/O operations, parsing and other preprocessing tasks. \\

To use DelSAT as a core tool for probabilistic inference comparable in usage with Tuffy, it was called by a small driver tool which translated rules and facts given in answer set syntax and any annotations of rules or facts with probabilities into DelSAT input format (aspif format (ASP intermediate language) and cost functions). The driver also invoked a small query tool which computed the probabilities of the query atoms (see further below). Reported times are the overall times including times spent by the driver tool, DelSAT and query tool unless where noted in the graph.\\

 Times have been averaged over five trials per experiment on a i7-4810MQ machine (4 cores, 2.8GHz) with 16GB RAM. DelSAT was compiled using Scala 2.12.7 and run with an OpenJDK 11. DelSAT was used in version 0.2.1 with default settings, except for the respective threshold and argument -mse (which activates optimized treatment of MSE cost functions, as this is the type of cost function used with probabilistic inference). For the Clingo-based tasks we have used Clingo 5.2.2 with Python 2.7.10 for scripting. Tuffy was used with its default settings and run using an OpenJDK 11 \footnote{(*) With our ``Smokers'' experiment Tuffy 0.4 performed on our machine massively better with an OpenJDK Java 11 JVM compared to other JVM versions which we had tried before with Tuffy, such as an OpenJDK 12 EA, for unknown reasons.}. \\
 
Cut-off graphs indicate timeouts. Some graphs start with small offset on the x-axis because the generated MLN or answer set programs required a certain minimum number of entities.\\

Where we could directly control accuracy (i.e., everywhere except Tuffy), experiments had been performed with different accuracy thresholds $\psi$.  The tolerance 20 specified with the reification-based tasks with 400 models corresponds roughly to accuracy $\psi \approx .001$ for coins and $\psi \approx 0.05$ for smokers.\\

In the coin game, a number of coins are tossed and the game is won if a certain subset of all coins comes up with ``heads''. The inference task is the approximation of the winning probability, calculated by counting the models with the winning coin combinations within the resulting sample and normalizing this count with the size of the sample. In addition, another random subset of coins are magically dependent from each other and one of the coins is biased (probability of ``heads'' is 0.6). This scenario contains probabilistically independent as well as mutually dependent uncertain facts. Also, inference difficulty clearly scales with the number of coins. Query atoms are \texttt{win}, \verb#coin_out(1,heads)# and \verb#coin_out(2,heads)# (recall that queries do not have any influence on the time required by DelSAT but only on the subsequent query computing task).\\
 In pseudo-syntax, such a randomly generated program looks, e.g., as follows:
{\vspace{-0.2cm} \scriptsize \begin{verbatim}
coin(1..8).
0.6: coin_out(1,heads).
0.5: coin_out(N,heads) :- coin(N), N != 1.
1{coin_out(N,heads), coin_out(N,tails)}1 :- coin(N).        
win :- 2{coin_out(3,heads),coin_out(4,heads)}2.
coin_out(4,heads) :- coin_out(6,heads).
\end{verbatim} \vspace{-0.2cm}}

The proposed methods cannot directly work with non-ground rules, so weighted non-ground rules everywhere have been translated into sets of ground rules, each (respectively, their corresponding auxiliary ``shortcut'' parameter atoms) annotated with the respective weights. \\
In ``Friends \& Smokers'', a randomly chosen number of persons are friends, a randomly chosen subset of all people smoke, there is a certain probability for being stressed ({\small \verb#0.3: stress(X)#}), it is assumed that stress leads to smoking ({\scriptsize \verb#smokes(X) :- stress(X)#}), and that friends influence each other with a certain probability 
({\scriptsize \verb#0.2: influences(X,Y)#}), in particular with regard to smoking:
 {\scriptsize \verb#smokes(X) :- friend(X,Y),# \verb#influences(Y,X), smokes(Y)#}. Smoking may leads to asthma ({\scriptsize \verb#0.4: h(X). asthma(X) :- smokes(X), h(X)#}). The query atoms are the ground atoms {\small \verb#asthma(X)#} per each person X.  \\
 
In a third experiment, we generate and sample from random graphs (note that these random graphs are different from those in an experiment with the same name in \cite{Nickles2018a}. To avoid the occasional generation of inconsistent programs, the integrity constraints have been omitted. On the other hand, all edges are now annotated with random probabilities, not just a subset).  Results are shown in Fig. \ref{fig:RandomGraphs}. The figure only shows results for Tuffy and DelSAT (the other approaches have been omitted here as they are clearly much slower), but it additionally shows the CPU time for DelSAT, i.e., the time required for mere sampling.\\
Random entities are here vertices of a directed graph (set $V$). Each edge between two different vertices has (as a logical fact) a random probability in $]0;0.3]$. Paths are construed using the groundings of the following rules {\footnotesize \vspace{-0.2cm} \begin{verbatim}
	path(X,Y) :- edge(X,Y), X != Y.
	path(Z,X) :- edge(Y,X), path(Z,Y),Y != Z.
	\end{verbatim} \vspace{-0.2cm}}
Query atoms are the paths $\{\mathit{Path} (v_1, v_i) | i \in V  \}$. \\
 This task has a significantly larger number of plain rules than Smokers or Coins, and steeper duration increase in the number of entities, due to the combinatorial explosion of edge-induced paths. \\ 
 
Overall, these experimental results do not show a clear winner between DelSAT and Tuffy - results are better for Tuffy in the ``Smokers'' experiment and worse for ``RandomGraphs'' and ``Coins''. The other tested approaches are significantly behind and seem to be useful merely as a proof of concept. While the solution using Clingo with reified models cannot compete with the native approach Diff-CDNL-ASP/SAT, it is still surprisingly fast, which, together with the fact that there is virtually no difference between the 400 and 800 model variants (the curves almost cover each other), exemplifies the strong solving performance of Clingo. However, these results also seem to indicate (since both Clingo/clasp and Diff-CDNL-ASP/SAT are based on CDNL) that our Diff-CDNL-ASP/SAT implementation could probably be made even faster by further optimizing its code or by using, e.g., C/C++ or Rust. The alternative approach using propagators is clearly usable only for tiny problems, at least with the current implementation (e.g., the graph for Diff-ASP-ThProp with $\psi = 0.001$ is not even visible in Fig. \ref{fig:Coins} because it immediately exceeded the vertical range). 


\section{Related Work}
\label{sec:relatedWork}

 \cite{chakraborty-aaai14} proposes distribution-aware sampling for SAT with weight functions over entire models (which might be seen as an instance of our MSE-style variant of cost optimization, albeit technically approached very differently). Also related is the weighted satisfiability problem which aims at maximizing the sum of the given weights of satisfied clauses (e.g., using MaxWalkSAT  \cite{Kautz96ageneral}), which can be used for inference in Bayesian networks. PSAT (Probabilistic Boolean Satisfiability) \cite{Pretolani2005} and SSAT problem \cite{DBLP:journals/jcss/Papadimitriou85} tackle related but different problems or use different formula semantics compared to our framework. PSAT has a setting similar to ours but decides on whether assignments of probabilities to formulas are consistent (which we assume as a prerequisite for sampling). 
An envisaged major application case for our multi-model sampling approach is probabilistic logic (programming) and probabilistic deductive databases, in particular Probabilistic ASP (of which existing approaches include, e.g., \cite{DBLP:journals/corr/abs-0812-0659,DBLP:journals/iandc/NgS94,ca7613df8f3c4bc591342bc8391977ae,NicMil15}), but we expect other uses cases too, such as working with combinatorial and search problems with uncertain facts and rules.
In contrast to the common MC-SAT sampling approach to inference with Markov Logic Networks (MLNs) \cite{mln} (a form of slice sampling), our task has a different semantics and our sampler is not wrapped around a uniform sampler, and we allow to specify rule or clause probabilities directly. An interesting approach with combines MLN with logic programming under the stable model semantics and compiles programs directly into ASP (using weak constraints) is \cite{ca7613df8f3c4bc591342bc8391977ae}. Other than our approach, existing approaches to machine learning-based SAT solving (such as Learning Rate Branching Heuristic (LRB)), or the ``hybrid'' combination of gradient descent / neural network-based techniques or numerical optimization with SAT (such as \cite{DBLP:journals/corr/abs-1802-03685}) aim largely at an improvement of SAT solving itself (which is not our concern in this work), or enable single model optimization. However, in the context of nonmonotonic Probabilistic Inductive Logic Programming, gradient descent has been used for weight learning (e.g., \cite{corapi,DBLP:journals/corr/NicklesM14,DBLP:conf/kr/LeeW18}. Recent related works \cite{Cohen2016TensorLogAD,Serafini2016LogicTN,DBLP:conf/nips/YangYC17} also explore logic-based inference or learning using deep learning frameworks (such as TensorFlow), utilizing gradient-based learning and differentiable inference processes. 

\section{Conclusion}
\label{sec:conclusion}  
We have presented differentiation-based approaches to SAT and ASP multi-model computation which sample models in order to minimizes a custom cost function. Using customized cost functions and parameter atoms, and configurable with a user-specified convergence threshold (thus allowing to trade off accuracy against computation time), our overall approach can be instantiated for example as a tool for probabilistic logic programming or for distribution-aware sampling of satisfiability witnesses or stable models. Building on previous work \cite{Nickles2018b,Nickles2018a}, we have presented an approach using a steepest descent method (with a first implementation as open source ASP and SAT solver DelSAT), an algorithm which utilizes Clingo's propagators, and finally an approach (not differentiation-based) which maps the problem to plain answer set optimization. Experimental results indicate that DelSAT is comparable in performance with an established framework for inference with Markov Logic Networks and that a reduction of the MSE-instance of the multimodel optimization to a conventional ASP or single-model optimization task (using reification) can not compete with a direct implementation in the decision-making process of CDCL/CDNL. Planned work comprises further experiments and the determination of formal convergence criteria.

\bibliographystyle{splncs04}

\bibliography{multimodelSATASP}

\end{document}